\pdfoutput=1

\documentclass[a4paper,fleqn]{cas-dc}

\usepackage[authoryear]{natbib}
\usepackage{array}
\usepackage{subfigure}
\usepackage{amsmath}
\usepackage{indentfirst}
\usepackage{float}
\usepackage{hyperref}
\def\tsc#1{\csdef{#1}{\textsc{\lowercase{#1}}\xspace}}
\tsc{WGM}
\tsc{QE}
\tsc{EP}
\tsc{PMS}
\tsc{BEC}
\tsc{DE}
\newcommand{\rone}[1]{\textcolor{black}{#1}}

\begin{document}
\begin{sloppypar}
\let\WriteBookmarks\relax
\def\floatpagepagefraction{1}
\def\textpagefraction{.001}
\shorttitle{Hybrid Tensor Decomposition in Neural Network Compression}
\shortauthors{Bijiao Wu et~al.}

\title [mode = title]{Hybrid Tensor Decomposition in Neural Network Compression}
\author[1]{Bijiao Wu}
\cormark[1]
\fnmark[1]
\address[1]{School of Electronic and Information Engineering, Xi'an Jiaotong University, Xi'an 710049, China}

\author[1]{Dingheng Wang}
\cormark[1]
\fnmark[2]

\author[1]{Guangshe Zhao}
\cormark[2]
\fnmark[3]

\author[2]{Lei Deng}
\fnmark[4]
\address[2]{University of California, Santa Barbara, CA93106, USA}

\author[3]{Guoqi Li}
\cormark[2]
\fnmark[5]
\address[3]{Department of Precision Instrumentation, Center for Brain Inspired Computing Research and  Beijing Innovation Center for Future Chip, Tsinghua University, Beijing 100084, China}

\cortext[cor1]{These authors contributed equally to this work}
\cortext[cor2]{Corresponding authors}
\fntext[fn1]{wbj123@stu.xjtu.edu.cn}
\fntext[fn2]{wangdai11@stu.xjtu.edu.cn}
\fntext[fn3]{zhaogs@mail.xjtu.edu.cn}
\fntext[fn4]{leideng@ucsb.edu}
\fntext[fn5]{liguoqi@mail.tsinghua.edu.cn}

\begin{abstract}
Deep neural networks (DNNs) have enabled impressive breakthroughs in various artificial intelligence (AI) applications recently due to its capability of learning high-level features from big data. However, the current demand of DNNs for computational resources especially the storage consumption is growing due to that the increasing sizes of models are being required for more and more complicated applications. To address this problem, several tensor decomposition methods including tensor-train (TT) and tensor-ring (TR) have been applied to compress DNNs and shown considerable compression effectiveness. In this work, we introduce the hierarchical Tucker (HT), a classical \rone{but rarely-used} tensor decomposition method, to investigate its capability in neural network compression. We convert the weight matrices and convolutional kernels to both HT and TT formats for comparative study, since the latter is the most widely used decomposition method and the variant of HT. We further theoretically and experimentally discover that the HT format has better performance on compressing weight matrices, while the TT format is more suited for compressing convolutional kernels. Based on this phenomenon we propose a \rone{strategy of hybrid tensor decomposition} by combining TT and HT together to compress convolutional and fully connected parts separately and attain better accuracy than only using the TT or HT format on convolutional neural networks (CNNs). Our work illuminates the prospects of hybrid tensor decomposition for neural network compression.
\end{abstract}




\maketitle

\section{Introduction}

Deep neural networks (DNNs) have shown the  state-of-the-art performance on many \rone{issues such as} computer vision~\citep{Krizhevsky2012ImageNet,He2016Deep}, natural language processing~\citep{inproceedings,SutskeverSequence}, reinforcement learning~\citep{DRL} and many other multidisciplinary fields~\citep{Heess2015Memory}. These advanced techniques come largely from available datasets, developed algorithms and powerful CPU \& GPU devices. However, as the deep learning problems become more and more complex, huge data and super large-scale DNNs become inevitable. \rone{Meanwhile}, the expensive hardware cost and long processing time further complicate the deployment of DNNs on resource constrained devices.

Hence, a large number of works~\citep{survey}, mainly including quantization~\citep{a,Wu_2018_WAGE}, pruning~\citep{Srinivas2015Data,Zhu_2018_PruneOrNot}, knowledge distillation~\citep{D}, compact convolutional filters~\citep{WuSqueezeDet} and low-rank factorization~\citep{alex2015tensorizing}, have been investigated  to reduce the hardware requirements and the running time \rone{of the deployment and application of DNNs}. Among these principal approaches, low-rank factorization\rone{, which tensorizes and decomposes the weight matrices into a series of low-rank tensors with the tensor decomposition theory originated with \citet{Hitchcock1927The},} has \rone{specific} advantages because of its solid mathematical theory and concise implementation. The relationship between neural networks and the tensor theory provides theoretical support for DNNs on one hand \citep{Cohen_2016_HT}, and enables the theory to be effectively applied in practical problems on the other hand \citep{Cohen_2016_HT2}. Consequently, several low-rank decomposition methods for compressing DNNs have attracted attention of many researchers in recent years.

As two of the most classical decomposition methods, CANDECOMP/PARAFAC (CP)~\citep{Carroll1970Analysis} and Tucker decomposition~\citep{Tucker1966Some} are earlier used to accelerate and compress convolutional neural networks (CNNs)~\citep{kim2016compression,lebedev2014speedingup}. With the development of the tensor decomposition theory, some new methods have achieved better results in various applications. The tensor-train (TT), which is first introduced by \citet{article}, decomposes a high-dimensional tensor into a series of 3-dimensional tensors. \citet{alex2015tensorizing} show that TT  can reduce redundancy of DNNs by compressing the dense weight matrices \rone{in} fully connected (FC) layers. Based on this, \citet{garipov2016ultimate} extend TT to convolutional kernels to compress the whole CNNs. In recent years, researchers have attempted to improve the memory~\citep{HuangLTNN} and the energy efficiency~\citep{tie} of DNNs \rone{by} further \rone{boosting} the application of TT decomposition. By contrast, some researchers focus on \rone{some} other tensor decomposition methods for compressing DNNs with different architectures, especially recurrent neural networks (RNNs). For example, the block-term decomposition (BTD)~\citep{DeLathauwer_2008_InventBTD}, which decomposes a high-dimensional tensor to a sum of multiple blocks in Tucker format, greatly reduces the parameters of RNNs and improves their \rone{training effectiveness} \citep{ye2017learning}. As a \rone{derived format}, tensor-ring (TR)~\citep{Zhao2018LearningET} exceeds TT in performance on RNNs and achieves promising results for video classification \citep{PanCompressing}. \rone{The most considerable physical significance of all} of these decomposition methods mentioned above in actual \rone{is to achieve the} sparse representation of the redundant weights\rone{. Naturally,} the ability of each \rone{tensor decomposition} method to represent high-dimensional data determines the performance of \rone{the corresponding} compressed DNNs. \rone{Doubtlessly}, there are obvious differences \rone{among} the inner structures of different decomposition methods, which \rone{might} be \rone{one of the most important reasons} that affects \rone{the} expressive ability \rone{of compressed DNNs,} but \rone{till now} is rarely discussed by previous researchers.

In this work, inspired by above studies, the flexible structure of the hierarchical Tucker (HT) decomposition method~\citep{GrasedyckHierarchical}, which iteratively factorizes a tensor into two subtensors by using Tucker and in actual is the source of TT, has aroused our interest. Since factorizing a tensor each time has more than one way, \rone{e.g., a tensor \(\mathcal{A} \in \mathbb{R} ^{n_{1} \times n_{2} \times n_{3} \times n_{4}}\) whose modes could be factorized as \(n_{1}n_{2} \times n_{3}n_{4}\) or \(n_{1} \times n_{2}n_{3}n_{4}\),} there are certainly multiple specific patterns for any single tensor. Therefore, the strong ability of HT to express high-dimensional data is derived from its varying structures. In general, these structures are treelike that mainly include two frequently-used patterns, namely balanced tree and degenerate tree. \rone{Moreover, researchers}~\citep{GrasedyckAn} have suggested that the balanced form of HT may require a lower rank than TT which is actually the degenerate tree form of HT. Therefore, we \rone{mainly} focus on HT with balanced tree in compressing DNNs. In the \rone{rest} of this paper, we only refer HT to the balanced tree unless otherwise indicated.

\rone{In our specific practices}, we apply HT decomposition to \rone{compress} both fully connected (FC) layers and convolutional kernels. Moreover, we compare the compressed DNNs in HT format with \rone{those} in TT format under theoretical analyses and extensive experiments, and the characteristics of compression performance of \rone{these two kinds of formats} are also studied. An important \rone{discovery} is that the HT and TT format have respective advantages in compressing different structures of DNNs, i.e., the HT format gains better performance in compressing the weight matrices \rone{in} FC layers, while the TT format is more suited for compressing convolutional kernels. \rone{We analyze and infer that the} reason underling this phenomenon is that HT prefers the tensor with balanced \rone{lengths} of dimensions, while it is better for TT to compress the tensor with unbalanced \rone{dimensions}. Based on this we propose a hybrid compression strategy consisting of \rone{TT-decomposed} convolutional and \rone{HT-decomposed} FC layers to improve the overall compression performance. The results of our experiments verify the effectiveness of the proposed method.

The  main contributions of this work are summarized as follows:

\begin{itemize}
\item \rone{It is the first time for HT decomposition being applied to compress DNNs to the best of our knowledge, and we experimentally show that HT has comparable compression ability to TT.}

\item \rone{We claim that HT is more sensitive than TT to the unbalanced tensor shape during gradient descending based on the theoretical analysis of the training process.}

\item \rone{We propose a strategy of hybrid tensor decomposition, in which HT and TT are separately applied in FC layers and convolutional kernels, outperforming using either of HT or TT.}

\item \rone{Our work enlightens to select the best tensorized strategy for specific DNNs and even design new neural architectures, which makes a step forward in deploying DNNs on embedded devices.}
\end{itemize}

The rest of this paper is organized as follows. In Section 2, we mainly introduce DNNs in the HT format. In Section 3, we \rone{analyze} the different advantages of HT- and TT-\rone{decomposed} networks and propose \rone{a strategy of hybrid tensor decomposition} which is further verified by our experiments in Section 4. \rone{Section 5 discusses some complex and interesting phenomena of our experiments.} Finally, we give our conclusion in Section 6.

\section{DNNs in the HT Format}

In this section, we give three presentations \rone{of HT decomposition}, i.e., basic, normal and contracted forms, all of which can replace the original uncompressed weights and rewrite the mapping from input signals to output signals. We also introduce the relationship between HT and TT\rone{, of which the latter is derived from the former}.

\subsection{HT Decomposition}

\subsubsection{Basic Form}

The hierarchical tensor format is introduced by~\cite{HackbuschA}, and on this basis~\cite{GrasedyckHierarchical} propose the HT decomposition. For a tensor $\mathcal{A}$$\in$$\mathbb{R}^{{n_1} \times {n_2} \times ... \times {n_d}}$, we can divide the dimensions into two sets, i.e., $t=\{ {t_1},{t_2},...,{t_k}\}$ and $s=\{ {s_1},{s_2},...,{s_{d - k}}\}$, \rone{to produce a matricization of \(\mathcal{A}\) like}  $A^{(t)}\in\mathbb{R}^{{n_{{t_1}}}{n_{{t_2}}}...{n_{{t_k}}} \times {n_{{s_1}}}{n_{{s_2}}}...{n_{{s_{d - k}}}}}$. Similarly, set \(t\) can also be grouped into ${t_l}$ and ${t_v}$, \rone{to obtain other two matrices $A^{(t_l)}$ and $A^{(t_v)}$}. \rone{If we define their corresponding column basis matrices as $U_t$, $U_{t_l}$, $U_{t_v}$, we could have}~\citep{Kressner_2011_HT}:
\begin{equation}
{U_t} = ({U_{{t_l}}}
\otimes {U_{{t_v}}}){B_t},
\label{equ1}
\end{equation}
where $U_{t}\in\mathbb{R}^{{n_{{t_1}}}{n_{{t_2}}}...{n_{{t_k}}} \times{r_t}}$, $U_{t_l}\in\mathbb{R}^{{n_{{t_{l_1}}}}{n_{{t_{l_2}}}}...{n_{{t_{l_i}}}} \times{r_{t_l}}}$, $U_{t_v}\in\mathbb{R}^{{n_{{t_{v_1}}}}{n_{{t_{v_2}}}}...{n_{{t_{v_{k-i}}}}} \times{r_{t_v}}}$ are called truncated matrices, and $B_{t}\in\mathbb{R}^{{{r_{{t_l}}}{r_{{t_v}}}} \times{r_t}}$ is termed as transfer matrix. All the $r_t$, \(r_{t_{l}}\), \(r_{t_{v}}\) are referred to as ranks. Besides, the operator $\otimes$ represents Kronecker product.

\begin{figure*}
\centering
\includegraphics{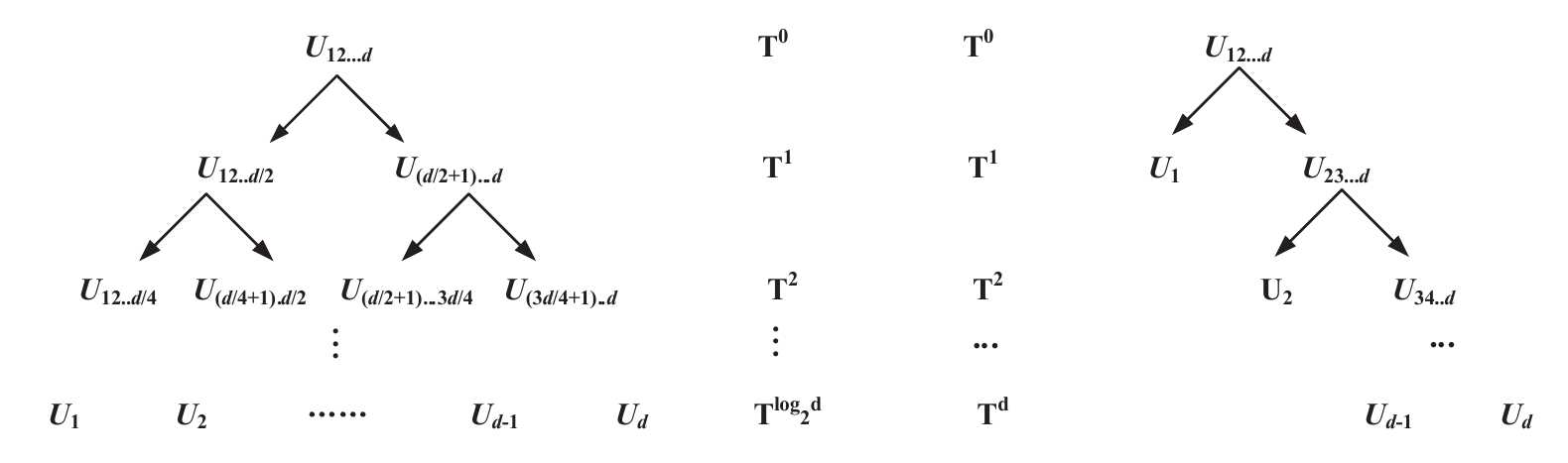}
\caption{Dimension tree. Left: balanced tree of HT. $d = {2^k}$ $(k = 1,2,3,...)$. Right: degenerate tree of TT.}
\label{fig:fig1}
\end{figure*}

By iteratively using Equation (\ref{equ1}) until all the rest truncated matrices cannot be further decomposed, the final structure of HT will come out. This process for \(\mathcal{A}\) can be generally expressed as follows:
\begin{align}
\begin{split}
\mathcal{A}\to
{U_{12...d}} =&  ({U_{12...{d \mathord{\left/
 {\vphantom {d 2}} \right.
 \kern-\nulldelimiterspace} 2}}} \otimes {U_{({d \mathord{\left/
 {\vphantom {d 2}} \right.
 \kern-\nulldelimiterspace} 2}+1)...d}}){B_{12...d}}\\
= & [(({U_{12...{d \mathord{\left/
 {\vphantom {d 4}} \right.
 \kern-\nulldelimiterspace} 4}}} \otimes {U_{({d \mathord{\left/
 {\vphantom {d 4}} \right.
 \kern-\nulldelimiterspace} 4}+1)...{d \mathord{\left/
 {\vphantom {d 2}} \right.
 \kern-\nulldelimiterspace} 2}}}){B_{12...{d \mathord{\left/
 {\vphantom {d 2}} \right.
 \kern-\nulldelimiterspace} 2}}})\\
\otimes& (({U_{({d \mathord{\left/
 {\vphantom {d 2}} \right.
 \kern-\nulldelimiterspace} 2}+1)...{{3d} \mathord{\left/
 {\vphantom {{3d} 4}} \right.
 \kern-\nulldelimiterspace} 4}}} \otimes {U_{({{3d} \mathord{\left/
 {\vphantom {{3d} 4}} \right.
 \kern-\nulldelimiterspace} 4}+1)...d}})\\
 &{B_{({d \mathord{\left/
 {\vphantom {d 2}} \right.
 \kern-\nulldelimiterspace} 2}+1)...d}})]{B_{12...d}}\\
 = &\cdots,\label{equ2}
\end{split}
\end{align}
where $U_{12...d}\in\mathbb{R}^{{n_{1}}{n_{2}}...{n_{d}} \times{1}}$ is reshaped from the tensor \(\mathcal{A}\) and the last ellipsis represents the subsequent decomposition of the tensor. \rone{Such presentation} is called the \emph{basic form} of the HT format, \rone{which can be drawn as a dimension tree} made up of $d$ truncated matrices and $d-1$ transfer matrices illustrated in Figure \ref{fig:fig1}. 

\subsubsection{Normal Form}

It is obvious that Equation (\ref{equ2}) is not very regular and simple to apply. In fact, it is \rone{easily observed} that when the HT process is finished, \(d\) truncated matrices will have only one corner mark, i.e., \(U_1, U_2,\cdots,U_d\). By using a common law ``\(AB \otimes CD = (A \otimes C)(B \otimes D)\)'', one can obtain a more concise HT format from Equation (\ref{equ2}) like \citep{Kressner_2011_HT}:
\begin{equation}
\begin{aligned}
U_{12 \cdots d} = &(U_1 \otimes U_2 \otimes \cdots \otimes U_d) \\
&(B_{12} \otimes B_{34} \otimes \cdots \otimes B_{(d-1)d}) \cdots B_{12 \cdots d},\label{eq_normal_ht}
\end{aligned}
\end{equation}
which is called the \emph{normal form} of HT since it is used more widely in tradition \citep{Hou_2015_HTApp,Kressner_2011_HT,Kressner_2014_HTMatlab}. The most remarkable characteristic of this form is that the calculation is arranged as level-to-level in the perspective of the dimension tree. \rone{Thus} one can directly design HT in the normal form for a tensor without iteratively calling Equation (\ref{equ1}).

\subsubsection{Contracted Form}

On the other hand, the so called dimension tree in Figure \ref{fig:fig1} is actually a kind of tensor network graph \citep{Espig_2011_TensorGraph}, since any two matrices in the dimension tree can be merged as one by using a contraction operator. In detail, if we reshape \(B_t \in \mathbb{R} ^{r_{t_l}r_{t_v} \times r_t}\) in Equation (\ref{equ1}) as a 3-dimensional tensor \(\mathcal{B}_{t} \in \mathbb{R} ^{r_{t_l} \times r_{t_v} \times r_t}\), we can have:
\begin{equation}
U_{t} = U_{t_l} \times ^{1} \mathcal{B}_{t} \times ^{1} U_{t_v},\label{equ1_contract}
\end{equation}
where $\times^1$ is called mode-1 contracted product \citep{Lee_2016_HTTT}. Then iteratively using this formulation, a new form to rewrite Equation (\ref{equ2}) \rone{or} (\ref{eq_normal_ht}) should be:
\begin{equation}
\begin{aligned}
U_{12 \cdots d} = &( \cdots ((U_1 \times ^{1} \mathcal{B}_{12} \times ^{1} U_2) \times ^{1} \mathcal{B}_{1234} \times ^{1} \\ & (U_3 \times ^{1} \mathcal{B}_{34} \times ^{1} U_4)) \times ^{1} \cdots \times ^{1} \\ & (U_{(d-1)} \times ^{1} \mathcal{B}_{(d-1)d} \times ^{1} U_d)\cdots),\label{eq_contract_ht}
\end{aligned}
\end{equation}
which is termed as the \emph{contracted form} of HT.

\begin{figure}
\centering
\includegraphics[width=8cm]{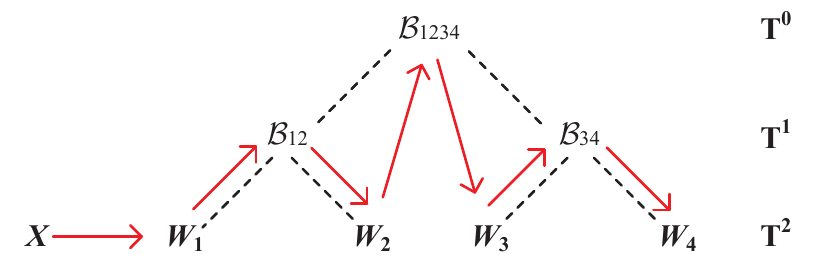}
\caption{The order of chain computation in a case of 4-dimensional tensor. The arrows represent the order of computation. $X$ represents the input data, $\mathcal{B}_i$ is the tensorized transfer matrix produced by each decomposition, and $W_i$ is the truncated matrix.}
\label{fig2}
\end{figure}

In this way, the expensive Kronecker product is avoided, thus the corresponding calculation process and probable middle results may not consume \rone{very} high computation and storage costs. For example, the contracted form of HT can allow an external input tensor to calculate with its truncated and transfer matrices one by one as shown in Figure \ref{fig2}, which appears to be the most efficient way that Equation (\ref{equ2}) and (\ref{eq_normal_ht}) cannot match.

\subsection{HT-Decomposed FC Layers in RNNs}

Note that FC layers \rone{with the weight matrix $W\in\mathbb{R}^{{M} \times{N}}$} are ubiquitous in DNNs \rone{especially RNNs}, \rone{and the parameters of FC layers account for a very high proportion of the total network parameters due to a large number of input and output neurons.} So the realization of compressing FC layers is of great significance for space complexity reduction. This section demonstrates the detailed \rone{process} to compress FC layers up to reconstruct the whole LSTM into the HT format.

\subsubsection{Tensorizing $x$ and $W$}

We tensorize the \rone{input} vector $x$ into a \(d\)-dimensional tensor $\mathcal{X}$ to obtain high-dimensional representation of the input data, while $W$ is transformed into a tensor with the same degree of dimensions \rone{for easier} HT decomposition. These tensorization operations are essentially regrouping the data into \rone{\(\mathcal{X}\in\mathbb{R}^{{n_1} \times {n_2} \times ...\times {n_d}}\) and \(\mathcal{W}\in\mathbb{R}^{{m_1n_1} \times {m_2n_2} \times ...\times {m_dn_d}}\) with the constraints $M = \prod\limits_{k = 1}^d {{m_k}} $ and $N = \prod\limits_{k = 1}^d {{n_k}} $ \citep{alex2015tensorizing}.}

\subsubsection{Compressing $\mathcal{W}$}

Following Equation (\ref{equ2}), (\ref{eq_normal_ht}) or (\ref{eq_contract_ht}), the \(d\)-dimensional tensor $\mathcal{W}$ derived from the original matrix can be easily represented as the HT format. It should be emphasized that, although the normal form defined in Equation (\ref{eq_normal_ht}) is more used in the past, for the tensorizing weight \(\mathcal{W}\), middle results produced by a serial of Kronecker products will cause heavy storage consumption, e.g., \(U_1 \otimes U_2 \otimes \cdots \otimes U_d\). This situation can even make the calculation process of compressed DNNs infeasible based on our practical experiences. \rone{Therefore}, we prefer the basic form in Equation (\ref{equ2}) and the contracted form in Equation (\ref{eq_contract_ht}) to compress weights in DNNs.

For the basic form, we reshape the transfer matrix $B_{12...d}$ and remove the Kronecker product of the first decomposition:
\begin{equation}
\begin{aligned}
&&{HT(\mathcal{W})} &={W_{1...{d \mathord{\left/
 {\vphantom {d 2}} \right.
 \kern-\nulldelimiterspace} 2}}}{B_{12...d}}{W{_{({d \mathord{\left/
 {\vphantom {d 2}} \right.
 \kern-\nulldelimiterspace} 2}+1)...d}}^T}\\
&&\ &=({W_{1...{d \mathord{\left/
 {\vphantom {d 4}} \right.
 \kern-\nulldelimiterspace} 4}}} \otimes {W_{({d \mathord{\left/
 {\vphantom {d 4}} \right.
 \kern-\nulldelimiterspace} 4}+1)...{d \mathord{\left/
 {\vphantom {d 2}} \right.
 \kern-\nulldelimiterspace} 2}}}){B_{12...{d \mathord{\left/
 {\vphantom {d 2}} \right.
 \kern-\nulldelimiterspace} 2}}}{B_{12...d}}\\
&&\ &B_{({d \mathord{\left/
 {\vphantom {d 2}} \right.
 \kern-\nulldelimiterspace} 2}+1)...d}^T{({W_{({d \mathord{\left/
 {\vphantom {d 2}} \right.
 \kern-\nulldelimiterspace} 2}+1)...{{3d} \mathord{\left/
 {\vphantom {{3d} 4}} \right.
 \kern-\nulldelimiterspace} 4}}} \otimes {W_{{({3d} \mathord{\left/
 {\vphantom {{3d} 4}} \right.
 \kern-\nulldelimiterspace} 4}+1)...d}})^T}\\
&&\ &=\cdots.\label{equ5}
\end{aligned}
\end{equation}
This process can significantly reduce the sizes of middle results, i.e., the result of Kronecker product of each two truncated matrices shall not be very large and it can be further reduced by contracting with the adjacent transfer matrix. For the contracted form, there is no need to preprocess some matrices, just to imitate Equation (\ref{eq_contract_ht}), we can have:
\begin{equation}
\begin{aligned}
HT(\mathcal{W}) = &( \cdots ((W_1 \times ^{1} \mathcal{B}_{12} \times ^{1} W_2) \times ^{1} \mathcal{B}_{1234} \times ^{1} \\ & (W_3 \times ^{1} \mathcal{B}_{34} \times ^{1} W_4)) \times ^{1} \cdots \times ^{1} \\ & (W_{(d-1)} \times ^{1} \mathcal{B}_{(d-1)d} \times ^{1} W_d)\cdots).\label{eq_weight_contract}
\end{aligned}
\end{equation}
According to Equation (\ref{equ5})-(\ref{eq_weight_contract}) and Figure \ref{fig:fig1}, \rone{the relevant space complexity of weight can be heavily reduced from \(\mathcal{O}({(mn)^d})\) to \(\mathcal{O}(dmnr+(d-1)r^3)\).}

\subsubsection{Computation between $\mathcal{X}$ and HT Matrices}

In the forward pass, the output of the HT-decomposed FC layer is obtained by the computation between the tensorized input $\mathcal{X}$ and the stored decomposed matrices, i.e., \(W_t\) and \(B_t\). In general there are two approaches to obtain the output, which are termed as \emph{recovery computation} and \emph{chain computation} respectively. The recovery computation is to recover the HT matrices to the corresponding original $\mathcal{W}$, and then multiply it by $\mathcal{X}$. The computation complexity of the recovery computation can be concluded as $\rone{\mathcal{O}}(({{\log _2}d}- 1)MN(r^3+r^2))$ by using Equation (\ref{equ5}).

Contrarily, the chain computation calculates the input with each truncated or transfer matrix in Equation (\ref{eq_weight_contract}) sequentially, because the operation between \(\mathcal{X}\) and \(\mathcal{W}\) is also contraction so that the appropriate order of contractions can be observed in the view of tensor network\rone{, e.g., Figure \ref{fig2}, where} we perform the operations between the input and matrices in the so called \rone{\emph{inorder traversal}}, \rone{i.e.,} each time the input should be reshaped into a befitting matrix to satisfy its modes with the matrices to be contracted. The computation complexity of the chain computation is expressed as $\rone{\mathcal{O}((2d - 1)n{\,}\max \{ M,N\}}{r^{1 + {{\log _2}d}}})$. In contrast, we find that the chain computation takes less time and is more suited for the forward pass, \rone{as} the expensive $M${\!}$N$ is erased. Specifically, according to Equation (\ref{eq_weight_contract}), we can roughly rewrite the product $W${\!}$x$ into the following form:
\begin{equation}
    \phi (W,x) = \mathcal{X}\times^1{W_1}\times^1{\mathcal{B}_{12}}\times^1{W_2}\times^1{\mathcal{B}_{1234}}\times^1\cdots\times^1{W_d},\label{chain}
\end{equation}
where \(\phi (\cdot)\) means it is not a strict representation of $W${\!}$x$ since the corresponding reshaping and transposing are omitted.


\subsubsection{HT-Decomposed LSTM Model}

Based on Equation (\ref{chain}), all kinds of FC layers in DNNs can be replaced with HT-decomposed \rone{ones} for network compression. For example, long-short term memory (LSTM)~\citep{HochreiterLong} is one of the advanced variants of RNNs, and its corresponding HT-decomposed version can be described as:
\begin{align}
\begin{split}
{\Gamma _u} &= \sigma (\phi({W_u},{x_t}) + \phi({R_u},{a_{t - 1}}) + {b_u}),\\
{\Gamma _f} &= \sigma (\phi({W_f},{x_t}) + \phi({R_f},{a_{t - 1}}) + {b_f}),\\
{\Gamma _o} &= \sigma (\phi({W_o},{x_t}) + \phi({R_o},{a_{t - 1}}) + {b_o}),\\
{\widetilde C_t} &= \tanh (\phi({W_c},{x_t}) + \phi({R_c},{a_{t - 1}}) + {b_c}),\\
{C_t} &= {\Gamma _u} \odot {\widetilde C_t} + {\Gamma _f} \odot {C_{t - 1}},\\
{a_t} &= {\Gamma _o} \odot \tanh ({C_t}),\label{equ9}
\end{split}
\end{align}
where $\sigma$ and $\tanh$ represent the sigmoid and hyperbolic function respectively, and $W_\theta$ (\(\theta=u,f,o,c\)) denotes the weight from the input to the hidden layer while $R_\theta$ presents the weight from the previous state to the next one.

\subsection{HT-Decomposed Convolutional Kernels in CNNs}

The convolutional kernel \(\mathcal{K}\) is the most crucial component of CNNs\rone{, and is} harder to compress because of \rone{its} inherent compactness. That is, parameters in \(\mathcal{K}\) are shared for different local fields of the input, \rone{and} each parameter should be responsible for many different data resources. \rone{Even worse,} this situation will be more severe if we compress the convolutional kernels to make their quantity of parameters lower. Nonetheless, CNNs still need to be compressed for wider applications, since the modern CNNs appear to be very deep, e.g., ResNet-152 \citep{He2016Deep}, so that the limited redundancy in convolutional kernels can be remitted to some extent in the view of the \rone{entire} CNN \citep{Denil_2013_Redundancy}.

\rone{In 2D-CNNs,} the kernel $\mathcal{K}\in\mathbb{R}^{l \times l \times C\times S}$ \rone{might} be decomposed in the HT format directly based on Figure \ref{fig:fig1}. However, further tensorization should also be considered for better compression performance through theoretical analyses and experimental verification~\citep{garipov2016ultimate}, i.e., the convolutional kernel should be transformed into \rone{\(\mathcal{K}\in\mathbb{R}^{{l^2} \times {c_1s_1} \times {c_2s_2}\times...\times {c_{d-1}s_{d-1}}}\) with constraints $C = \prod\limits_{k = 1}^{d-1} {{c_k}} $ and $S = \prod\limits_{k = 1}^{d-1} {{s_k}} $.} Then the new tensor can be converted into the $d$ truncated matrices and $d-1$ transfer matrices according to Equation (\ref{equ5}).

\rone{It should be emphasized that, although} the chain computation in Equation (\ref{chain}) is faster than the recovery computation, the former can be hardly utilized in convolution since there is no associative law between convolution and contraction. Therefore, as the same as \citet{garipov2016ultimate}, we can just adopt the recovery computation for HT-decomposed convolutional kernels.

\subsection{Relationship with TT}

\begin{figure}
\centering
\includegraphics[width=8cm]{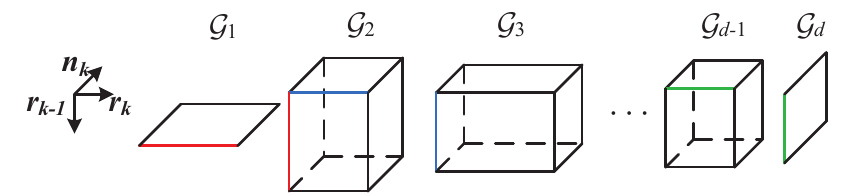}
\caption{TT cores for a $d$-dimensional tensor. The special case is ${r_0}={r_d}=1$. The same color represents the same dimensions which can be contracted.}
\label{figTT}
\end{figure}

\rone{For a tensor $\mathcal{A} \in \mathbb{R}^{{n_1} \times {n_2} \times ... \times {n_d}}$, \cite{article} proposes TT decomposition which has a more regular format as} \citep{Lee_2016_HTTT}:
\begin{equation}
   \mathcal{A} = {\mathcal{G}_1} \times^1{\mathcal{G}_2}\cdots\times^1 {\mathcal{G}_d},
\label{equ10}
\end{equation}
where $\mathcal{G}_{k}\in\mathbb{R}^{r_{k-1}\times n_{k}\times r_{k}}$ $({r_0}={r_d}=1)$ shown in Figure \ref{figTT} is termed as \emph{core tensor}. In fact, TT decomposition is a \rone{specific} variant of HT decomposition \citep{GrasedyckHierarchical,GrasedyckAn,Lee_2016_HTTT}. For example, in Equation (\ref{equ1}), set $t_l$ will have only one element if \(\mathcal{A}\) is in TT decomposition, and the following relationship should exist:
\begin{equation}
    \mathcal{G}_{k} = reshape({U_{t_l}}{B_t}),\label{equ11}
\end{equation}

\noindent where $U_{t_v}$ will be kept on decomposing according to Equation (\ref{equ1}). This process is shown in the right side of Figure \ref{fig:fig1} that presents as a degenerate tree, i.e., the TT format is the result of asymmetric HT decomposition and distinguishes from the HT format by the removal of Kronecker products and the formation of a uniform sequence of cores.

\section{Hybrid Compression}

In this section, we present and analyze the gradients of compressed DNNs in the HT and TT formats \rone{respectively}. We conclude that the DNN in the HT format should achieve better performance than that in TT when the dimensions are balanced, and thus HT is more suited for compressing FC layers whose lengths of the dimensions \rone{could} be balanced by more easy adjustment. However, we also discover that HT is more sensitive for unbalanced dimensions during training. \rone{Therefore}, we propose a hybrid compression method that combines HT and TT together to \rone{separately} compress FC layers and convolutional kernels of CNNs.

\subsection{Gradient Analyses}

\subsubsection{Basic Definition}

Error back propagation is usually applied in the training of DNNs. It is used to update the parameters by calculating the gradients which can be gained by the derivative of the error with respect to parameters. Thus, to discover the speciality of HT and TT in network compression, we focus on the gradients of compressed weights. For convenience, we uniformly use \(W_k\) to denote the truncated matrix \(U_k\) in HT or the matricized core tensor \(\mathcal{G}_k\) in TT.

In \rone{tensor-decomposed} DNNs, the gradient \rone{of the terminal \(k\)th compressed weight \(W_k\)} can be computed by the chain rule \rone{like}:
\begin{equation}
    {\frac{{\partial L}}{{\partial W_k}} } =  {\frac{{\partial L}}{{\partial y}}}{\frac{{\partial y}}{{\partial {W}}}}{\frac{{\partial {W}}}{{\partial W_k}}}, \label{equ14}
\end{equation}
where $L$ is the loss function, $y$ is the output, \rone{and} $W$ denotes the uncompressed weight matrix. Apparently, different decomposition methods bring about the main differences on the third term in Equation (\ref{equ14}), i.e., \(\partial W / \partial W_k\). Besides, the descending distance of each updating depends on the product between gradient value and learning rate. For compressed matrices with the same initialization, in general the gradient with larger size can get more value which results in a longer descending distance of the matrix, \rone{so we} also care about the size of the gradients.

\subsubsection{Gradient of HT}

\begin{figure}
\includegraphics[width=8cm]{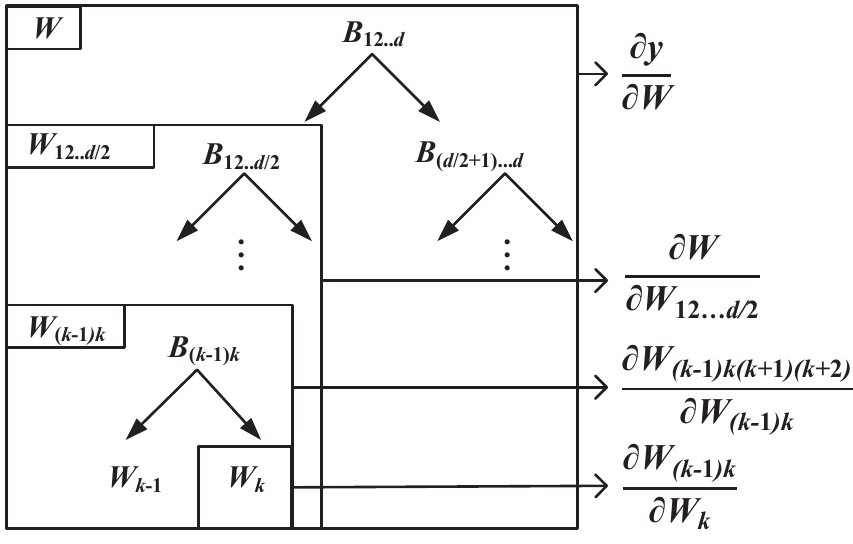}
\caption{Partial derivatives in dimension tree. Each rectangle is regarded as a whole.}
\label{figgra}
\end{figure}

It is known that in the HT dimension tree, any non-leaf node satisfies the relationship in Equation (\ref{equ1_contract}). For example, each corresponding minimum subtree at the last level in Figure \ref{fig:fig1} (left side) or Figure \ref{fig2} can be represented as:
\begin{equation}
    {W_{(k - 1)k}} = {W_{k - 1}}{B_{(k - 1)k}}{W_k},\label{W1}
\end{equation}
where the neighbouring elements $k-1$ and $k$ ($k=2,4,6,\dots ,d$) belong to the dimensions of the minimum subtree. Accordingly, the gradient \rone{of} each truncated matrix can be represented as:
\begin{equation}
\left\{
\begin{aligned}
\begin{split}
    &{\frac{{\partial {W_{(k-1)k}}}}{\partial W_{k-1}}} = [{B_{(k-1)k}}{W{_{k}}}] \otimes {I_n},\\
 &{\frac{{\partial {W_{(k-1)k}}}}{{\partial W_{k}}}} = {I_m} \otimes [{{W_{k-1}}{B_{(k-1)k}}}]^T,
\end{split}
\end{aligned}
\right.\label{equ17}
\end{equation}
where $n$ and $m$ represent the row of $W_{k-1}$ and $W_{k}$ respectively. 
We adopt the chain rule shown in Figure \ref{figgra} to compute the gradient of the uncompressed matrix $W$ with respect to each HT truncated matrix $W_k$: 
\begin{equation}
    \frac{{\partial W}}{{\partial {W_k}}} = \frac{{\partial W}}{{\partial {W_{1..k..d/2}}}}\frac{{\partial {W_{1..k..d/2}}}}{{\partial {W_{1..k..d/4}}}}...\frac{{\partial {W_{(k - 1)k}}}}{{\partial {W_k}}},\label{W2}
\end{equation}
where the number of partial derivatives is determined by the level of the dimension tree and reshaping is necessary for matrices due to the matricization of transfer tensors. For instance, for a three-level HT dimension tree, the gradient of $W_1$ should be:
\begin{equation}
    \frac{{\partial W}}{{\partial {W_1}}} = \frac{{\partial W}}{{\partial {W_{12}}}}\frac{{\partial {W_{12}}}}{{\partial {W_1}}},\label{eqw1}
\end{equation}
where the dimension tree satisfies ${W} = {W_{12}}{B_{1234}}{W_{34}}$ and ${W_{12}} = {W_{1}}{B_{12}}{W_{2}}$. The gradient in Equation (\ref{eqw1}) can be derived from the rules \rone{in} Equation (\ref{equ17})\rone{, then} the size of the gradient in Equation (\ref{W2}) is:
\begin{equation}
    size\left(\frac{{\partial {W}}}{{\partial {W_k}}}\right)=r_{k} \times {n_1}{n_2}\dots{n_{k-1}}{n_{k+1}\dots{n_d}}.\label{W4}
\end{equation}

\subsubsection{Gradient of TT}

For the network in the TT format, \cite{alex2015tensorizing} derive the gradient of the output with respect to the core matrices (slices of the core tensors) of the TT-\rone{decomposed} weights. To be more intuitive, we transfer the tensors to matrices and ignore the reshaping according to Equation (\ref{equ10})\rone{, then} the gradients of TT cores \rone{should} be governed by:

\begin{equation}
\left\{
\begin{aligned}
\begin{split}
\frac{{\partial {W}}}{{\partial {W_1}}} &= ({W_2}{W_3}...{W_d}) \otimes {I_n},\\
\frac{{\partial {W}}}{{\partial {W_k}}} &= ({W_{k + 1}}{W_{k + 2}}...{W_d}) \otimes {({W_1}...{W_{k - 2}}{W_{k - 1}})^T},\\
\frac{{\partial {W}}}{{\partial {W_d}}} &= {I_m} \otimes {({W_1}{W_2}...{W_{d - 1}})^T},
\end{split}
\end{aligned}
\right.\label{equ18}
\end{equation}

\noindent where $k=2,3,...,d-1$, $n=n_1$, and $m=n_d$. The sizes of these gradients are:
\begin{equation}
\left\{
\begin{aligned}
\begin{split}
size\left(\frac{{\partial {W}}}{{\partial {W_1}}}\right) &= {n_1}{r_1}\times{n_1}{n_2}...{n_d},\\
size\left(\frac{{\partial {W}}}{{\partial {W_k}}}\right) &={r_{k-1}}{r_k}\times{n_1}...{n_{k-1}{n_{k+1}}...{n_d}},\\
size\left(\frac{{\partial {W}}}{{\partial {W_d}}}\right) &={n_d}{r_{d-1}}\times{n_1}{n_2}...{n_d}.
\end{split}
\end{aligned}
\right.\label{equ19}
\end{equation}
According to Equation (\ref{W4}) and (\ref{equ19}), it is obvious that there is a certain difference between the gradients of HT and TT. The gradients of HT matrices have a unified form, while in TT the gradients are variable. The size of each gradient is \rone{largely} related to dimensional lengths $n$ \rone{that} may have different effects on updating the matrix in HT and TT\rone{, which will be analyzed concretely in the following.}

\subsection{Gradient Transfer of Varying Dimensions}\label{reason}

Previous experiments~\citep{alex2015tensorizing} show that better compression performance can be achieved by tensorizing weight matrices into tensors with balanced length of dimension\rone{s}. However, due to the fixed size of the filter, e.g., the most widely used \(3 \times 3\) filter, the unbalanced dimensions will occur in the \rone{process of tensorization of convolutional kernels}. Hence, in the following, we discuss two cases of HT and TT, i.e., equal and unequal length of dimensions, \rone{to} analyze the effect on the performance of compressed DNNs based on gradient transfer.

\subsubsection{Equal Length of Dimensions}

We assume that there is a tensor $\mathcal{A}$$\in$$\mathbb{R}^{{n_1} \times {n_2} \times ...{n_{d - 1}} \times {n_d}}$ that satisfies $n_{1}=n_{2}=\cdots=n_{d}=n$, and its every HT or TT rank equals $n$, which is also called mode-1 truncated rank produced by the matricization of \(A^{(\{1\})}\) or \(A^{(\{d\})}\) \citep{alex2015tensorizing,wang2019lossless}. Therefore, for DNNs in the HT format, the size of each compressed truncated matrix shall be $n\times n$. However, matricized core tensors in the TT format are different because \rone{$r_{0}=r_{d}=1$}, which means marginal core matrices $W_1$ and $W_d$ have a different amount of information. This situation causes the gradients related to $W_1$ and $W_d$ in TT-DNNs to be larger than \rone{those of} other core matrices at the second dimension, i.e., \(n^2 \times n^d\) in $W_1$ and $W_d$ vs. \(n^2 \times n^{d-1}\) in other matrices. \rone{To sum up, the} unbalanced distribution of information and different updating rate of gradients \rone{might} cause that the DNNs in the TT format have a lower precision than those in the HT format under the balanced dimension tree. In addition, this principle can also explain that TR, which is the same as TT except $r_{0}=r_{d}$ $\neq 1$, could achieve better compression performance than TT~\citep{PanCompressing,Zhao2018LearningET}.

\subsubsection{Unequal Length of Dimensions}

\begin{figure}
\centering
\subfigure[HT Network]{
\begin{minipage}{0.4\textwidth}
\centering
\includegraphics{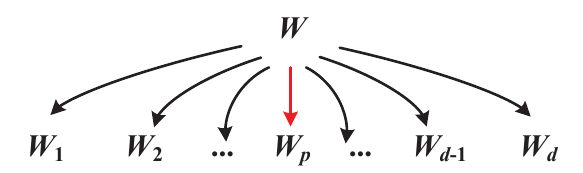} \\
\label{fig:fig2a}
\end{minipage}
}

\subfigure[TT Network]{
\begin{minipage}{0.4\textwidth}
\centering
\includegraphics{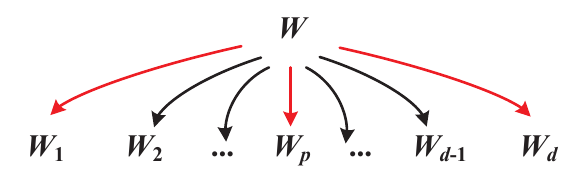} \\
\label{fig:fig2b}
\end{minipage}
}
\caption{Gradient transfer in the compressed network when the unequal length occurs. $n_p$ is smaller than others. The red arrow represents the larger value of the gradients in the gradient transfer.}
\end{figure}

However, when the length of the \(p\)th dimension $n_p$ of the tensor $\mathcal{A}$ greatly decreases, gradients of compressed weights in HT and TT formats will have different changes. For HT, according to Equation (\ref{W4}), the size of \rone{${\partial W}/{\partial {W_p}}$} is $r_{p} \times {n_1}{n_2}\dots{n_{p-1}}{n_{p+1}\dots{n_d}}$, which is larger than that of any other gradient due to the smaller term $n_p$. Then it brings a different updating rate of HT-truncated matrices with gradient transfer shown in Figure \ref{fig:fig2a}, where the value of gradient of the truncated matrix $W_p$ is larger than others. As for TT, the sizes of the first and last gradients are also larger than others except the \(p\)th gradient \rone{${\partial W}/{\partial {W_p}}$} according to Equation (\ref{equ19}), and the gradient transfer \rone{in} Figure \ref{fig:fig2b} \rone{illustrates that} more counts of larger gradients can be found \rone{compared} with HT. Thus, it is apparent that the effects on the final performance of HT and TT should be different. For HT, unequal \rone{dimensions} result in the consequence that the smaller matrix has a longer descending distance than others, while other matrices may update slower. The extremely unbalanced updating may make HT difficult to converge to the optimal solution and lead to poor performance. As for TT, different descending distances also exist, but there are more counts of matrices \rone{could} update faster which makes TT more robust to unequal \rone{dimensions compared} with HT, especially in the case of low degree of dimensions.

\begin{figure*}
    \centering
    \includegraphics[width=13cm]{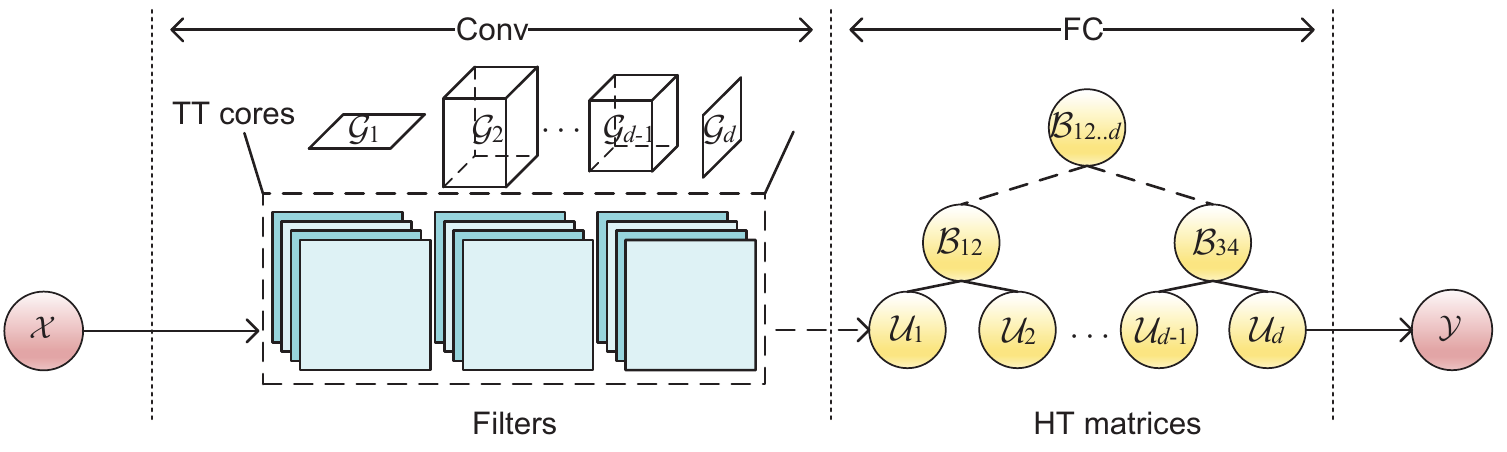}
    \caption{\rone{Illustration of the hybrid tensor decomposition of CNNs. The convolutional kernels are stored in the TT format and the dotted line surrounding the filters indicates the recovery computation of TT cores. The weight matrices of fully-connected layers are stored in the HT format.}}
    \label{fig:fighy}
\end{figure*}

\subsection{Hybrid \rone{Tensor Decomposition} Strategy}

The analyses above show that HT is more suited for compressing the tensor with balanced length of dimensions, while it is better for TT to compress the tensor with unbalanced dimensions. In practice, one can obtain the \rone{tensor with balanced} dimensions more easily in FC layers \citep{alex2015tensorizing} rather than convolutional kernels since the latter generally have the fixed filter size \citep{garipov2016ultimate}. For a network made up of different structures such as CNN, it is naturally not a good idea to compress the whole network with a single \rone{tensor decomposition} method, either HT or TT. To this end, we propose a hybrid \rone{tensor decomposition} strategy combining TT and HT together for CNNs consisting of convolutional kernels and FC layers. As depicted in Figure \ref{fig:fighy}, HT and TT are respectively responsible for compressing their suitable structures, specifically convolutional kernels are \rone{decomposed} in the TT format and weight matrices are stored in the HT format. We expect that the hybrid \rone{tensor decomposition} may achieve better performance than \rone{only using the single method HT or TT}.

\section{Experiments}

\rone{In this section, we will first aim at verifying the performance of HT-DNNs and comparing them with TT-DNNs to find whether HT is better for the weights with balanced dimensions. Then we emphasize the proposed hybrid tensor decomposition to compress 2D- and 3D-CNNs and check whether the hybrid strategy is better than either of HT or TT.}

\subsection{HT \rone{and TT Decomposition} on RNNs and CNNs}\label{4.2}

In the following we will make a wide comparison between HT- and TT-DNNs, including RNNs and CNNs. Recently, some tensor decomposition practices have achieved outstanding performance on video classification such as TT-LSTM~\citep{YangTensor} and TR-LSTM~\citep{PanCompressing}. For this reason, we also choose LSTM as the reference network to classify UCF11~\citep{Liu2009Recognizing} and UCF50~\citep{ReddyRecognizing}, which are two challenging video datasets. After that, we focus on the compression of convolutional kernels in CNNs based on \rone{the} CIFAR-10 dataset~\citep{KrizhevskyLearning} and adopt 3D-CNNs to recognize videos on UCF11 and CVRR-HANDS 3D datasets \citep{Ohn2014Hand}. We \rone{use stochastic gradient descent (SGD) to} train all of our networks from scratch with 0.9 momentum, and the learning rate is decayed 10 times per 30 epochs. \rone{This combination of hyper-parameters is a general selection in training process with SGD to some extent~\citep{alex2015tensorizing,learingrate,He2016Deep}.}

\subsubsection{Datasets and Training \rone{Details}}

\begin{figure}
    \centering
    \includegraphics[width=7cm,height=0.47\textwidth]{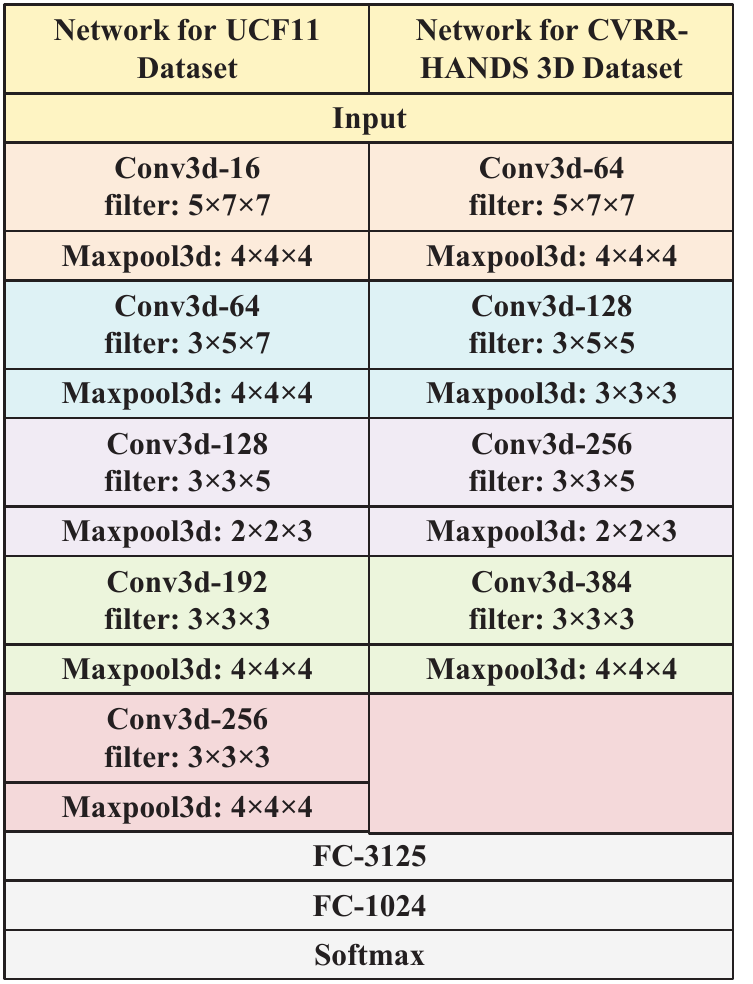}
    \caption{Network configuration. Left: network for UCF11 dataset. Right: network for CVRR-HANDS 3D dataset.}
    \label{fig:fig4}
\end{figure}

\paragraph{\textbf{UCF11 Dataset}} is also known as YouTube Action Data Set, which has 1600 video clips in total \rone{and includes} 11 action categories. \rone{Each} category is divided into 25 groups based on some common features, such as the same actor, similar background, similar viewpoint, and so on. Large variations \rone{of this dataset} make it very challenging because of the \rone{different} camera motion, object scale, viewpoint, etc. The dataset is processed differently for RNNs and CNNs in our experiments. In the experiments of compressing LSTM, we follow~\cite{YangTensor} to randomly sample 6 RGB frames \rone{with} size $160\times120$ from each clip \rone{and perform a 5-fold cross validation with mutual exclusive data splits}. We use the LSTM which contains a hidden layer with 2304 neurons\rone{, and for tensorization,} the input dimensions at each frame $160\times120\times3$ are reshaped to $15\times16\times16\times15$ and those of the hidden layer are $8\times6\times6\times8$. The initial learning rate is 0.001. For the experiments of compressing 3D-CNNs, we randomly extract 50 continuous RGB and optical flow frames \rone{with} size $80\times60$ from each clip. Then the stacked RGB and optical flow frames are fed into the two stream CNNs \rone{and their tensorized forms}~\citep{wang2019lossless}, which is shown on the left side of Figure \ref{fig:fig4}. The initial learning rate is still 0.001. \rone{However, we follow \cite{wang2019lossless} to select the leave one group to replace 5-fold cross validation since 3D-CNN might be more easily influenced by the similar background in the video clips of UCF11.}

\paragraph{\textbf{UCF50 Dataset}} consists of 6681 video clips collected from YouTube. Here we just try the same LSTM that is also used on UCF11 dataset\rone{, and the training details are still the same} without variation.

\paragraph{\textbf{CIFAR-10 Dataset}}\label{cvr} has a total number of 60000 $32\times32$ colour images \rone{which can be split into} 50000 training and 10000 validation samples, and is divided into 10 classes. We use VGG-19~\citep{Simonyan2014Very}, \rone{which is one of the best-performing CNNs on image recognition, as the reference model.} We compress the convolutional kernels in the HT and TT formats respectively with the same ranks \rone{as 16, and the dimensions of tensorization depend on the channels, e.g., \(64=4\times4\times4\), \(128=4\times8\times4\), \(256=4\times8\times8\), and \(512=8\times8\times8\)}. The initial learning rate is 0.01.

\paragraph{\textbf{CVRR-HANDS 3D (CVRR.) Dataset}} is also known as VIVA challenge's hand gesture dataset which consists of 886 intensity and depth hand gesture videos. It is sampled by using a Kinect under real-world driving settings to study natural human activities, and 19 types of gestures performing 8 subjects inside a vehicle are collected. Following the existing practice \citep{Molchanov_2015_3DCNN_1}, the dataset is expanded by mirroring and flipping in our experiment. For each video, we concatenate an intensity image, a depth image and a Sobel gradient image as \rone{a} \(28 \times 62 \times 3\) frame and make each video into uniform 32 frames as the final input. The \rone{referenced basic and compressed networks}~\citep{wang2019lossless} we used consist of four 3-dimensional convolutional layers and two FC layers and \rone{the} detailed architecture is shown on the right side of Figure \ref{fig:fig4}. Comparing with the settings given by~\citet{wang2019lossless}, \rone{we add} batch normalization after each FC layer. The initial learning rate is still 0.01.

\begin{table*}[width=1.8\linewidth,cols=6]
\caption{Experimental results on several datasets to verify the performance of HT\rone{-} and TT\rone{-decomposed} networks.}\label{tab:Table 7}
\begin{tabular*}{\tblwidth}{@{} LLrLrr@{} }
\toprule
Reference & Dataset & Baseline & Method  & Compr. & Accuracy \\
\midrule
\multirow{4}*{LSTM} &\multirow{2}*{UCF11}&\multirow{2}*{79.81\rone{$\pm$2.78}}                    &HT-LSTM   & 58.41 & \textbf{79.69\rone{$\pm$2.13}}    \\
                                           &&&TT-LSTM   & 56.32  & 76.12\rone{$\pm$1.60}      \\
\cline{2-6}
        &\multirow{2.5}*{UCF50}   &\multirow{2.5}*{77.93\rone{$\pm$1.02}} &\multirow{1.5}*{HT-LSTM} &\multirow{1.5}*{57.96}&\multirow{1.5}*{\textbf{76.56\rone{$\pm$0.63}}}      \\
                                           & & &\multirow{1.5}*{TT-LSTM}   & \multirow{1.5}*{55.91}  & \multirow{1.5}*{75.13\rone{$\pm$0.47}}      \\
\bottomrule
\multirow{6}*{CNN} &\multirow{2.5}*{UCF11} &\multirow{2.5}*{82.82\rone{$\pm$1.43}}&\multirow{1.5}*{HT-Conv} &  \multirow{1.5}*{8.49} &\multirow{1.5}*{81.50\rone{$\pm$2.01}}\\
                                            && &\multirow{1.25}*{TT-Conv} &    \multirow{1.25}*{8.49}&\multirow{1.25}*{\textbf{82.43\rone{$\pm$2.24}}}\\
\cline{2-6}
&\multirow{2.5}*{CIFAR-10} &\multirow{2.5}*{88.84\rone{$\pm$0.72}}&\multirow{1.5}*{HT-Conv} &\multirow{1.5}*{4.29}&\multirow{1.5}*{83.66\rone{$\pm$0.58}}\\
                                            && &\multirow{1.25}*{TT-Conv} & \multirow{1.25}*{4.29}&\multirow{1.25}*{\textbf{87.27\rone{$\pm$0.29}}}\\
\cline{2-6}
&\multirow{2.5}*{CVRR.} &\multirow{2.5}*{89.74\rone{$\pm$1.30}}&\multirow{1.5}*{HT-Conv}    & \multirow{1.5}*{7.37}  & \multirow{1.5}*{86.90\rone{$\pm$2.51}}      \\
                                                 && &\multirow{1.5}*{TT-Conv}  & \multirow{1.5}*{7.37}  & \multirow{1.5}*{\textbf{88.16\rone{$\pm$1.78}}}      \\
\bottomrule
\end{tabular*}
\end{table*}

\subsubsection{Results \rone{of} HT- and TT-\rone{decomposed} Networks }

To indicate the degree of compression, we term the ratio of uncompressed to compressed parameters of the same part as the compression factor (compr.).
\rone{Specifically for CNNs, we follow \citet{garipov2016ultimate} to indicate which part is compressed in TT or HT, e.g., `HT-Conv' means the convolutional kernels are compressed in HT format. To ensure the reliability of the experimental results, we carried out 5 times for each training setting. The mean and standard deviation of accuracy on these datasets are shown in Table \ref{tab:Table 7}.}

The experiments indicate there are different advantages in HT and TT for compression, which also evidence our earlier theoretical analyses of gradients. The performances of UCF11 and UCF50 datasets show that RNNs in the HT format have better performance than those in the TT format when compressing weight matrices, though they are a bit worse than uncompressed RNNs. For the results in CNNs where only convolutional kernels are compressed, it is clear that the performances of TT\rone{-decomposed} networks are closer to those of uncompressed networks, and HT\rone{-decomposed} networks get the worst score \rone{with obvious accuracy loss. Overall, obviously the compressed CNNs present lower compression factor, which might be related to the compact structure of the convolution kernel.}

\subsection{Hybrid \rone{Tensor Decomposition} on CNNs}

\begin{table*}[width=1.8\linewidth,cols=5]
\caption{\rone{Verification of the networks in hybrid tensor decomposition strategy.} }
\label{tab:Table 4}
\begin{tabular*}{\tblwidth}{@{} LrLrr@{} }
\toprule
Dataset & Baseline & Method  & Compr. & Accuracy \\
\midrule
 \multirow{3}*{UCF11} &\multirow{3}*{82.82\rone{$\pm$1.43}}
                                    &HT-Conv-HT-FC & 109.35 &79.83\rone{$\pm$1.77}       \\
                                    &&TT-Conv-TT-FC & 108.31 &78.98\rone{$\pm$1.43}        \\
                                    &&Hybrid (ours) & 106.26 &\textbf{80.60\rone{$\pm$1.90}}        \\
\midrule
 \multirow{3}*{CIFAR-10} &\multirow{3}*{88.84\rone{$\pm$0.72}}
                                    &HT-Conv-HT-FC & 34.17 &84.18\rone{$\pm$0.30}       \\
                                    &&TT-Conv-TT-FC & 34.65 &87.02\rone{$\pm$0.29}        \\
                                    &&Hybrid (ours) & 36.18 &\textbf{87.75\rone{$\pm$0.37}}        \\
\midrule
 \multirow{3}*{CVRR.} &\multirow{3}*{89.74\rone{$\pm$1.30}}
                                    &HT-Conv-HT-FC & 188.35 &89.61\rone{$\pm$1.34}       \\
                                    &&TT-Conv-TT-FC & 179.30 &89.16\rone{$\pm$1.63}        \\
                                    &&Hybrid (ours) & 171.38 &\textbf{90.00\rone{$\pm$1.70}}        \\

\midrule
 \multirow{3}*{ImageNet} &\multirow{3}*{58.56\rone{$\pm0.62$}}
                                    &HT-Conv-HT-FC & 12.67 &51.63\rone{$\pm0.67$}       \\
                                    &&TT-Conv-TT-FC & 11.61 &54.59\rone{$\pm0.28$}        \\
                                    &&Hybrid (ours) & 11.61 &\textbf{55.01\rone{$\pm0.13$}}        \\
\bottomrule
\end{tabular*}
\end{table*}

\rone{The above experiments present the different performances in HT and TT compression of fully connected and convolutional layers. In the following, we severally adopt HT, TT and hybrid tensor decomposition for compressing the whole CNNs.}

\subsubsection{\rone{Datasets and Training Settings}}

\paragraph{\textbf{Primary Verification.}} \rone{Our primary experimental verification is executed on the UCF11, CIFAR-10 and CVRR-HANDS 3D datasets following the previous training settings introduced in Section \ref{4.2}. The configuration of hyper-parameters of the fully connected layers and the convolutional layers of the hybrid decomposed networks are the same as the HT- and TT-\rone{decomposed} networks, respectively.}

\paragraph{\textbf{Task on ImageNet.}} To the best of our knowledge, there are few works considering tensor\rone{-decomposed} CNNs on large-scale datasets. In this experiment,  we do further experiments on ImageNet 2012 dataset~\citep{ImageNet}, which roughly \rone{consists of} 1.2 million training and 50,000 validation images \rone{with 1000 classes}. The dataset was used in the competition called the ImageNet Large-Scale Visual Recognition Challenge (ILSVRC) and is still the focus of researchers nowadays. We choose the classical Alexnet~\citep{Krizhevsky2012ImageNet} as the reference and compress the whole network \rone{in} the HT, TT and hybrid formats, respectively. The resized training images with the size of $256\times256$ are randomly cropped into $224\times224$ ones. We use the mini-batch of 128 and train the networks up to 100 epochs. The learning rate and the weight decay are selected to follow \citet{Krizhevsky2012ImageNet}. We exclude the dropout for the compressed networks \rone{to prevent under-fitting, and adopt the $224\times224$ center-crop in validation}.

\subsubsection{\rone{Results of Hybrid Tensor-decomposed CNNs}}

\rone{Like before, we use `HT-Conv-HT-FC' to represent the CNN whose convolutional parts and fully connected parts are compressed in the HT format, and so does `TT-Conv-TT-FC'. We use `Hybrid' to represent the network whose convolutional kernels are compressed in the TT format and fully connected weights are compressed in the HT format. }

\begin{figure}
\centering
\subfigure[CIFAR-10]{\label{fig:fig3}\includegraphics[width=4cm]{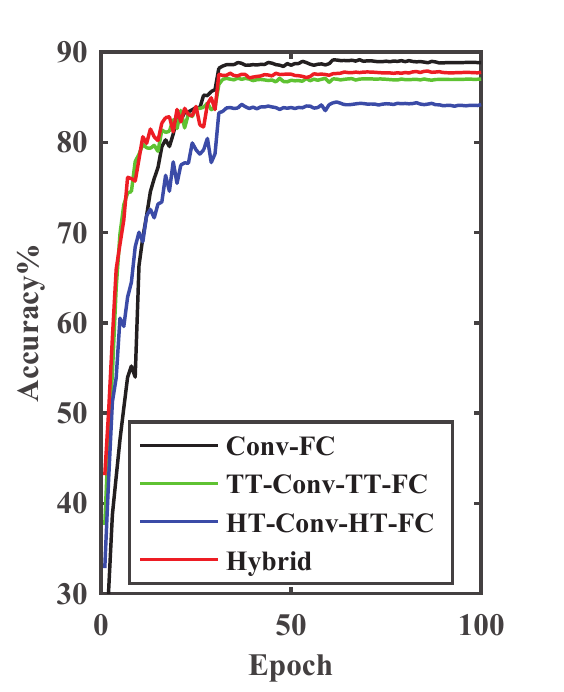}}
\subfigure[ImageNet]{\label{fig:figalex}\includegraphics[width=4cm]{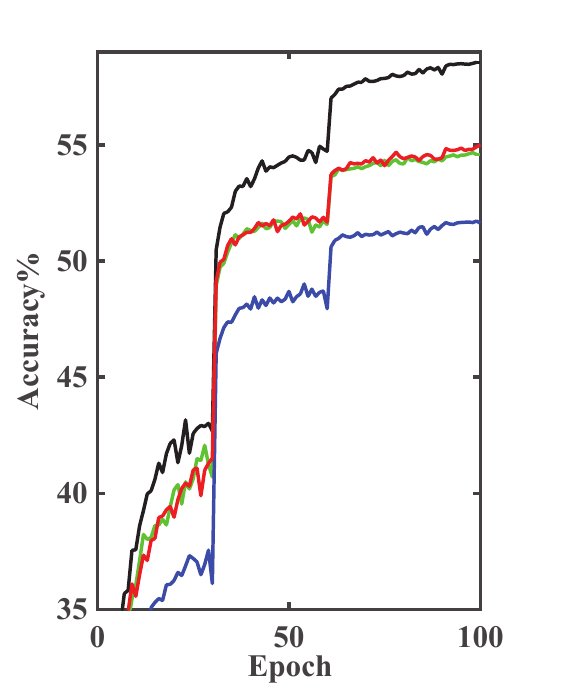}}%
\caption{Training curves on CIFAR-10 and ImageNet datasets.}
\label{f8}
\end{figure}

\rone{The results in Table \ref{tab:Table 4} verify that the hybrid CNNs can achieve the best classification results among compressed ones. Particularly, the hybrid 3D-CNN on CVRR. even surpasses the uncompressed one. For the results on the large-scale dataset, the network in the simplex HT format is inferior due to the large proportion of convolutional layers. By contrast, the hybrid CNN still has the best performance among the compressed CNNs. For CIFAR-10 and ImageNet datasets, we also illustrate their corresponding learning curves in Figure \ref{f8} as supplementary, which directly present the training processes of different networks and show the superior learning ability of hybrid CNNs. The learning ability of the HT-CNNs is obviously lagging behind other networks and the performance of hybrid CNNs improves on the basis of the TT-decomposed network.}

\section{\rone{Discussions}}

\rone{More and more compressed networks with low-rank factorization have been proposed, each of which has its own unique advantages but cannot suit any demands. Hence, a hybrid strategy is proposed in this paper. In this section we make some further analyses with the results of our experiments and extend to discussing some possible understanding of compressed networks.}

\subsection{\rone{Variable Length of Dimensions of HT}}

\rone{The experiments in section \ref{4.2} present various performances of the HT-decomposed RNNs and CNNs, and imply that HT is more friendly to weight matrices because of the easier tensorization. In this subsection, we intuitively discuss the effect of variable length of dimensions in HT decomposition to verify and explain whether our gradient analyses in Section 3 are reasonable. We carefully design and train a network with two FC layers on MNIST dataset~\citep{Mnist} and resize the original images from $28\times28$ to $32\times32$ to facilitate the setting of experimental parameters. The network consists of two layers, in which the first weight matrix has the size of $1024\times 1024$ and the second has the size $1024\times10$. We only compress the first layer as a 4-dimensional tensor and train several times with different settings of dimensions and ranks. }

\begin{figure}
    \centering
    \includegraphics[width=7cm]{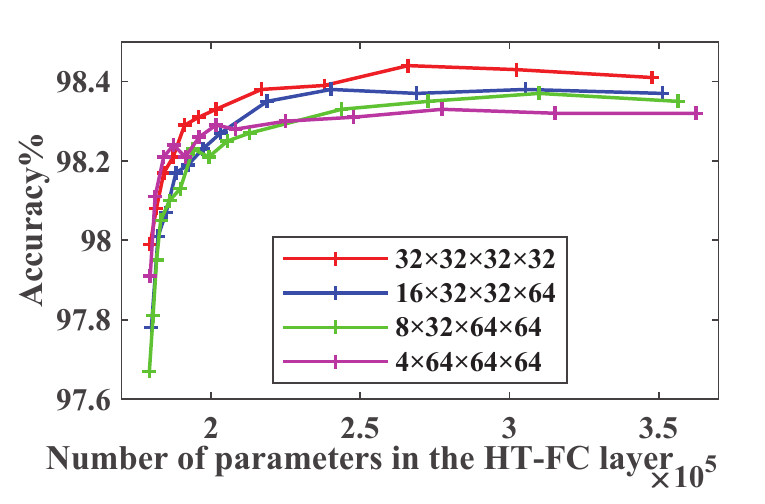}
    \caption{\rone{Experiment of examining the relationship between performance and shapes on the MNIST dataset. The legends represent different weight sizes of the HT-FC layer after tensorization.}}
    \label{mnist}
\end{figure}

\rone{The curves in Figure \ref{mnist} present the comparison between balanced dimensions and unbalanced dimensions of the HT-\rone{decomposed} network and indicate that the former style can get the best accuracy, thus our gradient analyses of HT is supported. Additionally, the compression factor of the balanced dimensions is higher than any other unbalanced dimensions. That is, when the same rank is taken, the parameters of the balanced dimensions will be fewer. Besides, the curves reflect that the performance of a compressed network will no longer improve when the number of parameters reaches a certain degree. Another point should be emphasized is that, in conditions of fewer parameters, there is only a slight difference of performances between the network with balanced dimensions and other designs. Therefore, choosing appropriate ranks is important to fully express the ability of network compression. Based on these experiences, we suggest to select the largest dimension length as the HT ranks, i.e., the largest mode-1 truncated rank.}

\subsection{\rone{Superiority of Hybrid Strategy}}

\rone{We have briefly introduced the superiority of hybrid strategy in the last section based on our experimental results. Here we would like to further explain some comprehensive phenomena to afford a more completed representation.}

\subsubsection{\rone{Variant Compr. under the Same Ranks}}

\rone{According to the experiments of LSTMs recorded in Table \ref{tab:Table 7}, we discover that different networks have various compression factors under the same ranking rule. HT-LSTM saves more memory under the same experimental settings compared with TT-LSTM, and the former even gets higher accuracy. This is completely accorded with the gradients analysis and corresponding experimental verification in Figure \ref{mnist}. Therefore, replacing the TT-FC layer with HT-FC layer in CNNs becomes a natural selection, and the experimental results in Table \ref{tab:Table 4} confirms this. On the other hand, since the HT is sensitive to unbalanced dimensions, HT-Conv appears to be an unadvisable choice.}

\subsubsection{\rone{Adaptability on Convolutional Kernels}}

\rone{According to Table \ref{tab:Table 7}, it is undoubted that HT is sensitive to unbalanced dimensions so that HT-Conv is not a sensible choice, which is also consistent with our gradient analysis in Section 3. However, for 3D-CNNs, the results of HT-Conv-HT-FC even exceed those of TT-Conv-TT-FC on UCF11 and CVRR. datasets. The probable reason might be that the tensorizing shapes of 2D- and 3D-convolutional kernels are different, and more importantly, this difference has potential influence on gradient transfer.}

\begin{figure}
\centering
\subfigure[2D Convolutional Kernel]{
\begin{minipage}{0.4\textwidth}
\centering
\includegraphics{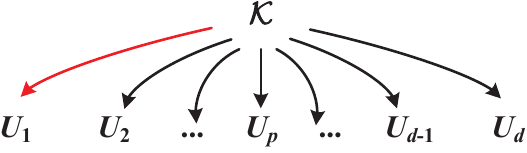} \\
\end{minipage}
}
\subfigure[3D Convolutional Kernel]{
\begin{minipage}{0.4\textwidth}
\centering
\includegraphics{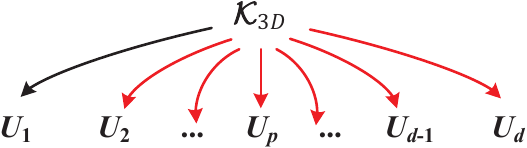} \\
\end{minipage}
}
\caption{\rone{Gradient transfer of 2D- and 3D-convolutional kernels in HT format. The red arrow represents the larger value of the gradients in the gradient transfer.}}
\label{fig:grad_2d_3d}
\end{figure}

\rone{In detail, the 3D-convolutional kernel has a cubic filter, thus the corresponding tensorization should be \(\mathcal{K}_{3D} \in \mathbb{R} ^{wht \times c_{1}s_{1} \times c_{2}s_{2} \times \cdots \times c_{d}s_{d}}\), where \(w\), \(h\) and \(t\) are filter sizes. Combining with the design of 3D-CNNs in Figure \ref{fig:fig4}, it can be noticed that usually there is \(wht > c_{i}s_{i}\) (\(i=1,2,\cdots,d\)). Contrarily, most 2D-convolutional kernels have the filter size of 3\(\times\)3, which often results in \(l^2 < c_{i}s_{i}\), since \(c_{i}\) or \(s_{i}\) is larger than 4 under most circumstances. According to the gradient transfer of HT illustrated in Figure \ref{fig:fig2a}, the larger portion of gradient from \(W\) will be attracted by the smaller mode \(l^{2}\), so many parameters in the compressed 2D-convolutional kernels could just be updated finitely. However, for 3D-convolutional kernels, smaller modes are all the \(c_{i}s_{i}\) so that the situation is opposite to 2D-convolutional kernels, then the gap between TT- and HT-decomposed 3D-CNNs is shrunk. Figure \ref{fig:grad_2d_3d} illustrates this difference vividly.}

\rone{Even so, HT still can not surpass TT in performance if we only compress convolutional kernels in 3D-CNNs. Nevertheless, combining with the advantages of HT-FC part, 3D-CNNs in pure HT format runs better than TT, let alone the hybrid CNNs.}

\subsubsection{\rone{Layer Coupling}}

\rone{Our results also indicate that compressing the whole CNN in hybrid tensor-decomposed format is better than only compressing convolutional kernels according to Table \ref{tab:Table 7} and \ref{tab:Table 4}. For example, when compressing two parts of CNN on the CIFAR-10 dataset, the accuracy can be improved from only compressing convolutional kernels even the former has fewer parameters. The reason is that there exists coupling between different parts of CNNs, i.e., compressing the entire CNN will not be much worse than only compressing convolutional or FC part \citep{wang2019lossless}. Another interesting phenomenon is that the CNN in HT format could surpass the one in TT format sometimes when the entire CNN is compressed, e.g., our results on UCF11 and CVRR. datasets, which is explained in the last subsection above.}




\subsection{\rone{Influence of Tensor Structure on Complexity}}

\rone{In this paper we mainly present the compressed networks with two treelike decomposition structures, i.e., HT and TT. Here we concretely discuss some differences of space and computation complexity among multiple decomposition methods but not limited to HT and TT.}

\rone{The stored format of parameters in compressed networks determines the space and computation complexity. We summarize the computation and space complexity of HT and TT in Table \ref{tblc}, and in addition, those of TR and BTD are also given for a comprehensive comparison. For the space complexity, all of these compression methods can greatly reduce the number of parameters in comparison with uncompressed networks. Due to $r^3$, the complexity of HT seems to be higher under the same and larger rank, but more information might be retained under the same amount of parameters because the rank required by HT could be smaller in this situation according to \citet{GrasedyckAn}. The large memory also occurs in BTD. However, TR has the most special space complexity since its tensorization is usually treated as a \(2d\)- rather than \(d\)-dimensional tensor \citep{Zhao2018LearningET,PanCompressing}, i.e., \(\mathcal{W} \in \mathbb{R} ^{m_1 \times m_2 \times \cdots \times m_d \times n_1 \times n_2 \times \cdots \times n_d}\), which is very effective to cut down the dimensional length. Generally speaking, for a fixed matrix \(W \in \mathbb{R} ^{M \times N}\), higher \(d\) results in lighter space complexity. When dimensional lengths are reduced to $mn/p$, the new dimension shall rise to $d{\log _{mn/p}}mn$. It is obvious that the product of dimensions $d{\log _{mn/p}}mn$ and dimensional lengths $mn/p$ has dropped, which leads to the lower space complexity.}

\rone{In the aspect of running time, the simple and unified format like the TT can greatly reduce the computation complexity due to the advantages of sequenced contractions through each core tensor in the forward pass \citep{alex2015tensorizing}. While the low correlation between the truncated matrices in HT (truncated matrices should be connected by transfer matrices) results in higher computation complexity, which is also more sensitive to rank. For TR whose computation complexity is close to HT, since its specific tensorizing strategy as we mentioned, the computation complexity is still higher than that of TT even the product $M${\!}$N$ is avoided. With respect to BTD, the computation complexity should be the highest because of its exponential sensitive to ranks.}

\rone{All in all, TT is derived from the HT degenerate tree, and has lower flexibility than HT but is simple to calculate in DNNs. Meanwhile, the advantage of HT is its potential of maintaining information under the same level of space complexity (ranks may be different). Besides, TR is a relatively particular case whose space complexity might be smaller but computation complexity is higher, and BTD has no advantage in terms of complexity.}

\begin{table*}[width=1.8\linewidth,cols=3]
\caption{\rone{Comparision among uncompressed FC layer, HT\rone{-}FC layer, TT\rone{-}FC layer, TR\rone{-}FC layer and BTD\rone{-}FC layer on forward computation and space complexity. The baseline is derived from the weight matrix \(W \in \mathbb{R} ^{M \times N}\), $N=n_1 \times n_2 \times \cdots \times n_d$, $n$ is the maximal \(n_k\)(\(k=1,2,\cdots,d\)), $M=m_1 \times m_2 \times \cdots \times m_d$, $m$ is the maximal \(m_k\)(\(k=1,2,\cdots,d\)), $r$ is the maximal rank and $C$ is the CP-rank.}}
\label{tblc}
\begin{tabular*}{\tblwidth}{@{} Lrr@{} }
\toprule
Method & Computation & Space \\
\midrule
FC&   $\rone{\mathcal{O}}(MN)$    &  $\rone{\mathcal{O}}(MN)$     \\
HT format& $\rone{\mathcal{O}((2d - 1)n{\,}\max \{ M,N\} r^{1 + {{\log _2}d}}})$ & $\rone{\mathcal{O}}(dmnr+(d-1)r^3)$    \\
TT format& $\rone{\mathcal{O}}(dn{\,}\max\{ M,N\} r^2)$ & $\rone{\mathcal{O}}((d-2)mn{r}^2+2mnr)$     \\
TR format& $\rone{\mathcal{O}}(d(M+N){r^3})$ & $\rone{\mathcal{O}}(d(m+n){r}^2)$     \\
BTD format& $\rone{\mathcal{O}}(dn{\,}\max\{ M,N\}{r^d}C)$ & $\rone{\mathcal{O}}((dmnr+r^d)C)$     \\
\bottomrule
\end{tabular*}
\end{table*}

\subsection{\rone{Deeper Understanding of Tensor-decomposed DNNs}}

\rone{In general, compressed networks are considered to be derived from uncompressed networks, and the generated accuracy loss usually exists. In our experiments, most of results show that the classification ability of the compressed network declines as well. However, the performance of the compressed network may far exceed that of uncompressed networks in some specific cases~\citep{PanCompressing,YangTensor,ye2017learning}, which appears to be very counter-intuitive. We change the the number of neurons of the hidden layer of LSTM in section \ref{4.2} from 2304 to 256 in the light of previous network configurations \citep{PanCompressing,YangTensor,ye2017learning} and train the uncompressed, TT-, HT-, TR- and BTD-LSTMs on UCF11 dataset successively. For the compressed networks except TR, the input dimensions are reshaped to $8\times20\times20\times18$ and the output dimensions are $4\times4\times4\times4$. The hyper-parameters of TR-LSTM follows \citep{PanCompressing} and other experimental settings are the same as \citep{YangTensor}.}

\begin{figure}
    \centering
    \includegraphics[width=8cm]{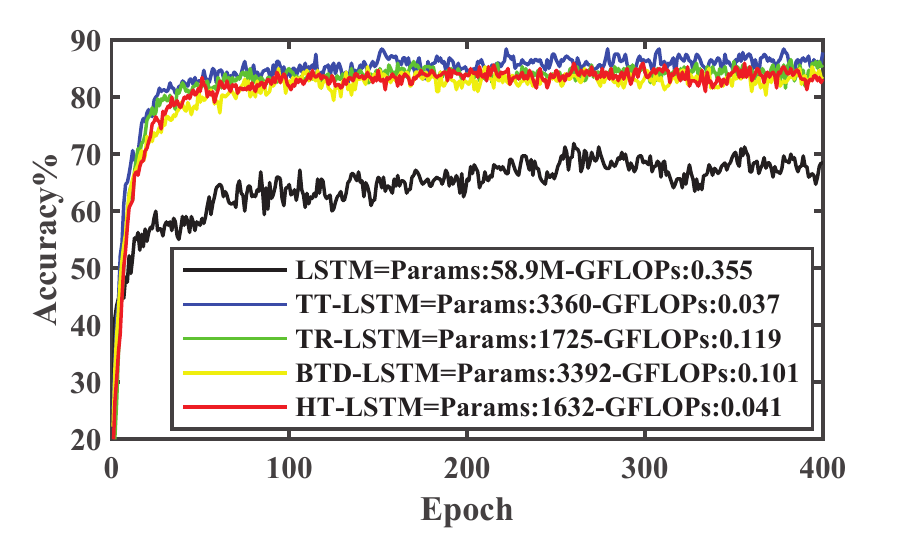}
    \caption{\rone{Performance of different LSTMs with 256 hidden neurons on the UCF11 dataset.}}
    \label{ucf11256}
\end{figure}

\rone{The corresponding best learning curves in Figure \ref{ucf11256} present the outstanding ability of compressed LSTMs obviously, i.e., the accuracy of any kind of compressed LSTM is greatly improved from the uncompressed one. Moreover, as we use the Adam optimizer here by following \citet{PanCompressing,YangTensor,ye2017learning}, the long-term performances fluctuate slightly, where each compressed LSTM is more stable than that of uncompressed one, although there are a little differences in these compressed LSTMs. Comparing the results in Table \ref{tab:Table 7}, the uncompressed LSTM here has poor fitting ability due to its less hidden neurons, however, all the compressed LSTMs achieve much better performance with fewer parameters and faster operations.}

\begin{figure}
    \centering
    \includegraphics[width=8cm]{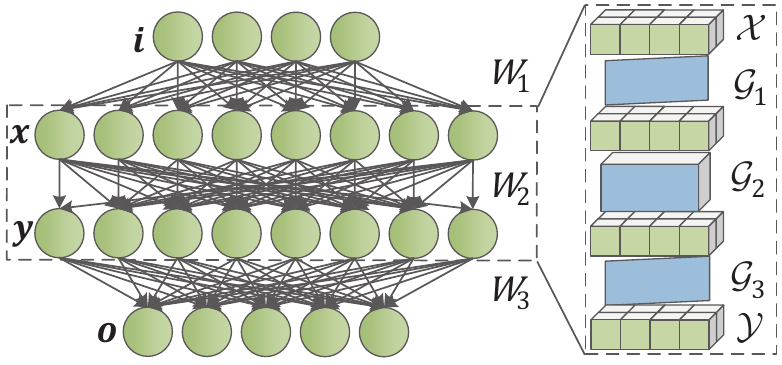}
    \caption{\rone{An example demonstrating   that the tensor-decomposed network can be regarded as a new network that is independent of the uncompressed network to some extent. Here a network  with three layers of weights could be restructured as a new one with five   layers of weights  by TT. Two layers of hidden neurons turn into four layers.}}
    \label{fig:tt_independent}
\end{figure}

\rone{We deem that all these results indicate that the tensor-decomposed network can be regarded as a new network that is independent of the original  uncompressed network to some extent. In  general the compressed DNN is hardly to surpass the relevant original one, unless redesign a new compact neural architecture, e.g., neural architecture search (NAS) \citep{survey}. That is to say, the tensor architecture has ability to reconstruct the topology of neurons, or even describe and explain the potential mechanism of DNNs \citep{Cichocki_2018_TensorNetworks}. Taking TT as an example because of its simplicity, a new and compact network could be constructed based on the original one as shown in Figure \ref{fig:tt_independent}, if we ignore that there is no nonlinear activation functions among the dotted box. In a sense, the tensor-decomposed DNN changes the original architecture of uncompressed network, since the new data flow is produced, particularly the chain computation of HT or the sequenced contractions of TT \citep{alex2015tensorizing} is utilized. By contrast, quantization and pruning do not have the similar characteristic, i.e., it is difficult for them to abandon their correlated uncompressed DNNs. Consequently, this situation implies that the tensor decomposition method might be perpendicular to quantization and pruning, and forebodes the promising joint-way compression based on these different methods in future.} 

\section{Conclusion}

In this paper, we apply HT decomposition with the balanced pattern to compress RNNs and CNNs for the first time, and propose a new hybrid compression strategy to improve the performance of tensor\rone{-decomposed} CNNs. With tens of times the compression factor, the accuracy is approximately maintained, which indicates that high redundancy in the original network can be weakened through the proposed method.

We also compare the HT format with the TT format that originates from the unbalanced pattern of HT decomposition. One important finding is that the HT format is more suited for compressing weight matrices and the TT format has advantages on compressing convolutional kernels. The proposed hybrid compression based on different advantages of HT and TT formats shows advanced overall performance. These results contribute to choosing the best strategy for neural network compression and make a step forward in deploying neural networks on embedded devices.

Last but not least, we would like to note that there are various other techniques having  been proposed to compress DNNs in recent years, such as data quantization~\citep{a,Wu_2018_WAGE}, network sparsification~\citep{SutskeverSequence,Zhu_2018_PruneOrNot} and neural architecture search~\citep{Barret2018Learning,ZophNeural}.  In this work, we only focus on  the hybrid tensor decomposition method by decomposing large-scale parameter representation into a series of smaller tensors
to shrink the memory volume and operation number. For the future work, we believe that the joint-way compression across multiple techniques to pursue an extreme compression factor is of great potential.

\section*{Acknowledgement}
This work is partially supported by National Key R\&D Program of China (No.2018YFE0200200), Beijing Academy of Artificial Intelligence (BAAI), and a grant from the Institute for Guo Qiang, Tsinghua university, and key scientific technological innovation research project by Ministry of Education, and the open project of Zhejiang laboratory.


\bibliographystyle{cas-model2-names}

\bibliography{reference}

\end{sloppypar}
\end{document}